# COLLABORATIVE KNOWLEDGE SHARING AND EDITING

Philippe MARTIN     *Université de La Réunion (and adjunct researcher of Griffith University, Australia)*
                    *EA2525 LIM, Saint-Denis de la Réunion, F-97490, France.*
pmji@phmartin.info

**ABSTRACT**

This article first lists reasons why - in the long term or when creating a new knowledge base (KB) for general knowledge sharing purposes - collaboratively building a well-organized KB does/can provide more possibilities, with on the whole no more costs, than the mainstream approach where knowledge creation and re-use involves searching, merging and creating (semi-)independent (relatively small) ontologies or semi-formal documents. The article lists elements required to achieve this and describes the main one: a KB editing protocol that keeps the KB free of automatically/manually detected inconsistencies while not forcing them to discuss or agree on terminology and beliefs nor requiring a selection committee.

**KEYWORDS**

Knowledge sharing, integration, retrieval and evaluation.

## 1. INTRODUCTION

Ontology repositories - and, more generally, the Semantic Web - are most often envisaged as composed of many small static (semi-)formal files (e.g., RDF or RDFa documents) more or less independently developed, hence loosely interconnected and with many *implicit* redundancies and inconsistencies between them (in this article, "implicit" means "not represented in formal or semi-formal way"). For example, this mainstream approach is advocated by Shadbolt et al. (2006) and Casanovas et al. (2007). The missing interconnections are difficult to recover manually and automatically. As Section 2 shows, due to these missing semantic relations, this mainstream "*static file* based approach" - as opposed to an approach based on *one* (distributed or not) "collaboratively-built well-organized large knowledge base (cbwoKB)" - makes knowledge re-use tasks complex to support and do correctly or efficiently, especially in a collaborative way. Most Semantic Web related research works are intended to support such tasks (ontology creation, retrieval, comparison and merging). However, most often, they lead people to create new files - thus contributing to the problems of knowledge re-use - instead of inserting their knowledge into one cbwoKB. Such a KB may be on a single machine or may be *a global virtual cbwoKB (gv-cbwoKB) distributed into various correlated cbwoKBs on several Web servers and/or people's machines of a peer-to-peer network*. To avoid *implicit* redundancies and inconsistencies within a gv-cbwoKB, there should be direct/indirect cross-references and knowledge assertion+query forwarding between the cbwoKBs. This point is not detailed in this article. Martin (2009) introduces a protocol to support this, based on having each cbwoKB i) defining and advertising the kinds of knowledge objects it stores, ii) committing to store all objects fulfilling this advertised definition, and for other objects, iii) pointing/redirecting to a relevant cbwoKB. This protocol is not detailed in this article.

   Except for WebKB-2 (www.webkb.org; Martin and Eboueya, 2008) - the tool implementing the new techniques described in this article - no other ontology/KB server has an ontology-based protocol permitting and enforcing or encouraging people to interconnect their knowledge into a *cbwoKB, while keeping it **at-least-***

***minimally-well-organized*** *(this means that manually or automatically detected partial redundancies or inconsistencies are prevented or made explicit via relations of specialization, identity and/or correction)* **and *without forcing people to agree*** *on terminology nor beliefs (knowledge integration is loss-less)*. Indeed, i) achieving these two requirements for scalable cooperative ontology building is often but wrongly assumed to be impossible or to involve centralization or domain restrictions, ii) it requires the users to see and write (semi-)formal knowledge representations, iii) it does not permit to directly re-use already existing ontologies, iv) it requires proposing and managing a large general ontology (WebKB-2 does so), and iv) it is useful for general repositories but then only indirectly for applications. In general repositories, as we shall see, choices between contradictory beliefs need not and should not be made. Thus, for each application performing problem-solving, its developers should make selections and perform choices based on the requirements of the application..

Other KB servers/editors (e.g., Ontolingua, OntoWeb, Ontosaurus, Freebase, CYC and semantic wiki servers) have no such protocols and i) let all/some users modify what other ones have entered (this discourages information entering or leads to edit wars), or ii) require all/some users to approve or not changes made in the KB, possibly via a workflow system (this is bothersome for the evaluators, may force them to make arbitrary selections, and this is a bottleneck in information sharing that often discourages information providers). By avoiding these two governance problems and leading to a well organized KBs, such kinds of cbwoKB protocol form a basis for a scalable knowledge sharing, even when multiple communities are involved. Actually, unlike with other approaches, a same cbwoKB can be used by many communities with partially overlapping focus since the KB is organized and can be filtered, queried or browsed by each person according to her needs or according to a community viewpoint. Even if built by many communities a (virtual) cbwoKB is unlikely to be huge since i) redundancies are reduced, and ii) "well organized knowledge" (as opposed to data) is difficult to build. However, a cbwoKB can permit to index or relate the content of data-bases. In any case, the bigger and the more organized the cbwoKB, the more information are easier to access and compare. Since building a cbwoKB can partly re-use resources of more classic (i.e., less organized) Semantic Web solutions or database solutions, it can be incrementally built to overcome the limitations of these solutions when they become clear and annoying to their users.

Section 3 presents the knowledge representation model used by the rules of the collaborative "KB editing" protocol of WebKB-2. Section 4 presents these rules and introduces many ideas yet unpublished in a journal. For readability reasons, the model and rules are not presented in a fully formal way. Furthermore, as with most methodological rules, the "completeness" criterion does not apply well to these rules.

Collaborative evaluation of knowledge representations is an extension of collaborative KB editing since, for precision and re-use purpose, evaluations should themselves be knowledge representations. The collaboration scheme of WebKB-2 is quickly introduced in Point 7 of the collaborative "KB editing" protocol.

WebKB-2 has been applied to the collaborative representation of many domains by students (for learning purposes), researchers (for knowledge sharing and evaluation purposes) and, currently, experts in biodiversity. Section 5 presents an example of application for (e-)learning.

Section 6 concludes and reminds that the presented knowledge sharing approaches are complementary.

## 2. APPROACHES BASED ON FILES VERSUS CBWOKB SERVERS

With files, information retrieval (IR) often leads to a list of possibly relevant files or *pieces of information* (**objects**, e.g., a formal term or a informal sentence) whereas it leads to an exact answer in one ontology (a cbwoKB or one formal file; the problem is that without a cbwoKB, there are more than one file). Such an answer may be a portion of the cbwoKB, e.g., a part/subtask/specialization hierarchy (with associated argumentation structures) if the query is of the kind "what are the resources/tools/methods to do ...". Such semantically structured answers allow a user to find and compare all relevant objects instead of getting a long redundant list of objects/files where original/precise ones are hidden among/behind objects that are more general, mainstream or from big organizations. This is also why IR quality decreases when the size and number of the files increases, but not when the number of objects increases in one ontology.

The more objects two files contain, the more difficult it is to link these files via semantic relations and hence to semantically compare, organize and evaluate them. Instead, similarity/distance (statistical) measures have to be used. In a cbwoKB, when needed, semantic queries can be used to filter objects or generate files, according to arbitrary complex combinations of criteria, e.g., about the creators of the objects. (Some of these criteria may be used for the internal organization of the cbwoKB but the resulting "views" or "contexts" are language/content dependent choices and, unlike (semi-)independently created static files, lead the users to strongly relate objects of different views). Ontology libraries, from early ones such as the Ontolingua library to imagined ones such as "The Lattice of Theories" (Sowa, 2005), are often organized into "minimal and internally consistent theories" to maximize their re-use. However, this also leads to few relations between objects of different ontologies, as well as implicit redundancies or inconsistencies between them, and hence more difficulties to compare, merge or relate them. On the other hand, as acknowledged by Sowa, if the objects are organized into a cbwoKB, such (lattices of) theories can be generated via queries.

With formal files as inputs and outputs, knowledge re-use or integration leads to the creation of even more files and requires people to select, compare, relate, merge, adapt and combine (parts of) files. Except for simple applications where fully automatic tools can deliver good-enough results, these are complex tasks that have to be done by trained people who know the domain. Most works in collaborative knowledge sharing or "ontology evolution in collaborative environments" are about (semi-)automatic procedures for integrating ontologies (Euzenat et al, 2009) and for rejecting or integrating changes made in other ontologies, e.g., (Casanovas et al, 2007; Noy and Tudorache, 2008; Palma et al, 2008). In a cbwoKB, no adaptation or integration has to be done for each re-use: the most important relations from an object have to be entered by its creators and then can be complemented by any user. Indeed, it is often the case that only the object creators know what their objects really mean or have information required for relating them to other objects.

The normalization/editing rules of a cbwoKB should maximize the use of principled multi-inheritance hierarchies (for example, hierarchies of specialization/mereological/spatial/... relations) where each object has a "right place" in the restricted sense that different users would search or insert a same object at the same place. Only a KB server with a large cbwoKB can permit a knowledge provider to simply/directly add one new object "at its right place" and guide her to provide precise and re-usable objects that complement the already stored objects. This "unique/right place", i.e., the absence of implicit redundancies, is a minimal requirement for knowledge insertion and retrieval to be done in a scalable way in the hierarchies and hence in the semantic network of which they are the backbones (Dromey, 2006).

## 3. LANGUAGE MODEL FOR THE KB EDITING PROTOCOL

The cbwoKB editing protocol used in WebKB-2 is intended to keep the cbwoKB "at-least-minimally-well-organized" in the sense given in the introduction. It is not tied to any particular knowledge representation language (KRL) or inference mechanism (hence, this is not the point of this article and no comparison is made on such mechanisms). This protocol only requires that *conflicts between knowledge representations - i.e., partial redundancies or inconsistencies between terms or statements* - are detected by some inference mechanism or by people (hence, the protocol also works with informal pieces of knowledge as long as they can be inter-related by semantic relations). This does not imply that the KR language should be restricted. The more conflicts are detected, the more the KB is kept organized and hence exploitable.

The *model for the protocols* - i.e., their view on a KB (whichever KR language it actually uses) - is a set of *objects* which are either *terms* or *statements*. Every object has at least one associated source (creator, believer, interpreter, source file or language) represented by a formal term. A *formal term* is a unique identifier for anything that can be though of, i.e., either a source, a statement or a category. It has a unique meaning which may be made partially/totally explicit by its creator via *definitions* with necessary *and/or* sufficient conditions. An *identifier* may be an URI or, if it is not a creator identifier, may include the identifier of its creator (this is the classic solution to avoid lexical conflicts between terms from various sources). An informal term is one name of one or several objects. Two objects may share one or several names but cannot share identifiers. A *statement* is a sentence that is either formal, semi-formal or informal. It is informal if it cannot be translated into a logic formula, for example because it does not have a formal grammar with an interpretation in some logics. Otherwise, it is formal if it only uses formal terms, and semi-formal if it uses some informal terms. A statement is either a *category definition* or a *belief*. A belief must have a source that is its creator and that believes in it and/or that has represented (and hence interpreted) a statement from some other source. Finally, a *category* is either a *type* of objects or an *individual* (object). A type (a "class" in OWL) is either a relation type or a concept type. An individual is an instance of a first-order type.

Giving a definition is equivalent to using a specialization/identity relation, except that the system can exploit the definition to better place the term in the specialization hierarchy. Every belief is also automatically inserted in the specialization hierarchy and its place may be refined by its creator if this does not introduce an inconsistency in the KB. In order to have a unique specialization/generalization hierarchy and hence be able to compare any pair of formal or informal objects (i.e., know if one generalizes or specializes the other), this hierarchy must actually use several kinds of specialization relations (all of which being subtypes of an "extended-specialization" relation type): i) the classic "subtype" and "instance" relations between formal terms, ii) the classic "logical-deduction-of" between formal statements (which, when formal terms have definitions, permits to calculate or check subtype/instance relations between these terms), and iii) an "informal-generalization" from a formal or informal object to an informal one.

The *KR model of WebKB-2*, its associated notations and its inference mechanism must now be introduced for illustration purposes. Although graph-based, this model is equivalent to the model of KIF (Knowledge Interchange Format; http://logic.stanford.edu/kif/dpans.html), i.e., it permits to use 1st order logic with collections (sets, lists, ...) and contexts (meta-statements that restrict statements). WebKB-2 allows the use of several notations: RDF/XML (an XML format for knowledge using the RDF model), the KIF standard notation and other ones which are here collectively called KRLX. These KRLX languages were specially designed to ease knowledge sharing: they are expressive, intuitive and normalizing, i.e., they guide users to represent things in ways that are automatically comparable. One of them is a formal controlled English named FE. It will be used for the examples along with KIF. These languages can be used for creating assertion/query commands and these commands can be sent to the WebKB-2 server via the HTTP/CGI protocol, from an application or from a WebKB-2 Web form. Other communication interfaces are being implemented: one based on SOAP and one based on OKBC (Open Knowledge Base Connectivity; http://www.ai.sri.com/~okbc) to query (or be queried by) frame-based tools or servers, e.g., Loom, SRI and the GKB-Editor.

Here are examples of terms in KRLX. `en#"bird"` and `"bird"` refer to the English informal word "bird" while `wn#bird` is a formal term referring to one of the WordNet categories for "bird". Here are examples of statements in FE. `u1#u2#"birds fly"` is an informal statement from u2 that is represented by u1. `u1#`any u1#bird is pm#agent of a pm#flight´` is a formal statement and definition by u1 of `u1#bird` as something that necessarily fly. `u1#`every u1#bird is agent of a flight´` is a semi-formal statement and belief of u1 that "every u1#bird flies". In KIF, these last two statements would respectively be `(creator u1 '(defrelation u1#bird (?b) :=> (exists ((?f pm#flight)) (pm#agent ?b ?f))))` and `(believer u1 '(forall ((?b u1#bird)) (exists ((?f flight)) (agent ?b ?f))))`.

When the creator of an object is not explicitly specified, WebKB-2 exploits its "default creator" related rules and variables to find this creator during the parsing. Similarly, unless already explicitly specified by the creator, WebKB-2 uses the "parsing date" for the creation date of a new object. The creator of a belief is also encouraged to add contextualizing relations on it (at least temporal and spatial relations must be specified).

RDF/XML - the W3C recommended linearisation of RDF - and OWL - the W3C recommended language ontology - are currently not particularly well suited for the cbwoKB editing protocol or, more generally, for the representation or interconnection of expressive statements from different users in a same KB.

- They offer no standard way to associate a believer, creator or interpreter to every object in an RDF/XML file. Since 2003, RDF/XML has no `bagID` keyword, thus no way to represent contexts and hence believers or beliefs. XML name-space prefixes (e.g., `u1:bird`), Dublin Core relations and statement reification do not permit to do this. This is likely a temporary only constraint since many RDF-related languages or systems extend RDF in this direction: Notation3 (N3), Sesame, Virtuoso, ...
- RDF and OWL - like almost all description logics - do not permit their users to distinguish definitions from universal quantifications. More precisely, they do not offer a universal quantifier. N3 does (Turtle, the RDF-restricted subset of N3, does not). The distinction is important since, as noted in the documentation of KIF (http://logic.stanford.edu/kif/dpans.html#5.3), a universally quantified statement (belief) may be false while a definition cannot. A definition may be said to be "neither true nor false" or "always true by definition". A user u1 is perfectly entitled to define `u1#cat` as a subtype of `wn#chair`; there is no inconsistency as long as the ways `u1#cat` is further defined or used respect the constraints associated with `wn#chair`. A definition may be changed by its creator but then the meaning of the defined term is changed rather than corrected. This distinction is important for a cbwoKB editing protocol since it leads to different conflict resolution strategies: "term cloning" and "loss-less correction" (Point 5 and Point 6 of the next section).

- Many natural language sentences are difficult to represent in RDF/XML+OWL or N3+OWL, since they do not yet have various kinds of numerical quantifiers, contexts, collections, modalities, ... (FE has concise syntactic sugar for the different kinds). However, at least N3 might soon be extended.
- Like most formal languages, RDF/XML and N3 do not accept - or have a special syntax for - the use of informal objects instead of formal objects. KRLX does and this permits WebKB-2 to create one specialization/generalization hierarchy categorizing all objects. More precisely, this is an "extended specialization/generalization" hierarchy since in WebKB-2 the classic "generalization" relation between formal objects (logical implication) has been extended to apply to informal objects too.

For its cbwoKB editing protocol, WebKB-2 detects *(partial) redundancies or inconsistencies between objects* by detecting exclusion and extended specialization relations between (parts of) these objects. A statement Y is an *extended specialization* of a statement X (i.e., Y includes the information of X and hence *either contradicts it or makes it redundant*) if X structurally matches a part of Y and if each of the terms in this part of Y is identical or an extended specialization of its counterpart term in X. For example, WebKB-2 can detect that `u2#`Tweety can be agent of a flight with duration at least 2.5 hour´` (which means "u2 believes that Tweety can fly for at least 2.5 hours") is an extended specialization (and an "extended instantiation") of both `u1#`every bird can be agent of a flight´` and `u1#`2 bird can be agent of a flight´`. In KIF, the first of these two statements can be written:
`(believer u1 '(modality possible '(forall ((?b bird)) (exists ((?f flight)) (agent ?b ?f)))))`

Furthermore, these last two statements can be found to be extended specializations of (and redundant with) *respectively* `u2#`75% of bird can be agent of a flight´` and `u2#`at least 1 bird can be agent of a flight´`. Similarly, this last graph can be found to be exclusive with `u3#`no bird can be agent of a flight´`.

WebKB-2 uses the same graph-matching technique for calculating partial or total extended-specialization relations between formal/informal statements, and therefore also "actual or potential conflicts". *Other inference mechanisms could be used* instead or in addition *for detecting more specialization relations*. This matching takes into account numerical quantifiers and measures, not just existential and universal quantifiers. Apart for this, it is similar to the classic graph matching for a specialization (or conversely, a generalization which is a logical deduction) between positive conjunctive existential formulas (with or without an associated positive context, i.e., a meta-statement that does not restrict its truth domain). This classic graph matching is sound and complete with respect to first-order logic and can be computed with polynomial complexity if the query graph (X in the above description) has no cycle (Chein and Mugnier, 1997). Apart from this restricted case, graph matching for detecting an extended specialization is not always sound and complete. However, this operation works with language of any complexity (it is not restricted to OWL or FOL) and the results of searches for extended specializations of a query graph are always "relevant".

## 4. COLLABORATIVE KB EDITING PROTOCOL

The rules of the protocol are intended for each object to be connected to at least another object via relations of specialization/generalization, identity and/or argumentation. These rules also permit a loss-less information integration since they do not force to make knowledge selections. They apply to the addition, modification or removal of an object in the KB, e.g., through a graphical interface or via the parsing of a new command in a new input file. This does not serialize objects in the KB and waiting till the whole input file is parsed would not permit to detect more *partial redundancies or inconsistencies between the objects*.

The independence of the protocol with respect to the KRL is clear in its high-level algorithms which are given below in Java (and, for clarity purposes, in an object-oriented way) and then discussed via a list of informal rules. These algorithms present some checks on a user's attempt to remove or add a statement and the resulting system decision: rejecting the action ("return false") or accepting it, with possibly some automatic repair step before accepting it. Only statement removal and adding are considered in the algorithms since i) updating is considered as removing followed by adding, ii) reading or re-using an object is always accepted (privacy control is not dealt with in this article), and iii) term removal or adding must be made via the removal or addition of a statement (see the second informal rule below). In the following algorithms and rules, the word "user" is used as a synonym for "source".

```
boolean statement.removal_by (User agent)
{ if (object.creator != agent) return false;
  if (agent.created_statements.are_inconsistent_with(this)) return false;
  if (agent.created_statements.are_redundant_with(this)) return false;
  if (this.is_definition())
  { if (KB.statements_without(this).are_inconsistent())
       KB.clone_term_in_statements_using(this.defined_term());
  }
  else if (KB.statements_without(this).are_inconsistent()) this.clone_for_other_believers();
  KB.remove(this,agent);   return true;
}

boolean statement.adding_by (User agent)
{ if (this.is_informal_statement() && !this.has_associated_argumentation_relation()) return false;
  if (agent.created_statements.are_inconsistent_with(this)) return false;
  if (agent.created_statements.are_redundant_with(this)) return false;
  if (this.is_definition())
  { if (this.is_definition_of_new_term() &&  KB.statements.are_inconsistent_with(this)) return false;
    if (this.is_new_definition_of_already_declared_term() &&
        KB.statements.are_inconsistent_with(this))
      KB.clone_term_in_statement_inconsistent_with(this);
  }
  else if (KB.statements.are_partially_conflicting_with(this))
        return false;   //"implicitly redundant/inconsistent"
  KB.add(this,agent);   return true;
}
```

Here are the informal rules enforced by these algorithms.

1. *Any user can add and use any object* but *an object may only be modified or removed by its creator*.

2. *Adding, modifying or removing a term* is done by adding, modifying or removing at least one statement (generally, one relation) that uses this term. *A new term can only be added by specializing another term* (e.g., via a definition), except for process types which, for convenience purposes, can also be added via subprocess/superprocess relations. In WebKB-2, every new statement is also automatically categorized into the extended specialization hierarchy. A new informal statement must also be connected via an argumentation relation to an already stored statement. In summary, all objects are manually or automatically inserted in the extended specialization hierarchy and/or the subprocess hierarchy, and thus can be easily searched and compared. However, it is clear that if one user (say, u2) enters a term (say, u2#object) that is implicitly semantically close to another user's term (say, u1#thing) but does not manually relates them or manages to give u2#object a definition that is not automatically comparable to the definition of u1#thing (i.e., there is no partial redundancies between the two definition) then the two terms cannot be automatically related by the system and the implicit redundancy cannot be rejected by the system. Here, the problem is that u2 has not respected the following "best practice" rule (which is part of WebKB-2 normalization rules): "always relate a term to all existing terms in the KB via the most important or common relations: i) transitive relations, especially (extended) specialization/generalization relations and mereological relations (to specify parts, containers, …), ii) exclusion/correction relations (especially via subtype partitions), iii) instance/type relations, iii) basic relations from/to processes, iv) contextualizing relations (spatial, temporal, modal, …) and v) argumentation relations".

3. *If adding, modifying or removing a **statement** introduces an **implicit redundancy** (detected by the system) in the shared KB, **or** if this introduces a detected **inconsistency between statements believed by the user***

*having done this action*, this **action is rejected** by the system. Thus, in the case of an addition, the user must refine his statement before trying to add it again or he must first modify at least one of his already entered statements. An "implicit" redundancy is a redundancy between two statements without a relation between them making the redundancy explicit. Such a relation is typically an equivalence relation in the case of total redundancy and an extended specialization relation (e.g., an "example" relation) in the case of partial redundancy. As illustrated in the previous section, the detection of extended specializations between two objects reveals an inconsistency or a total/partial redundancy. It is often not necessary to distinguish between these two cases to reject the newly entered object. Extended "instantiations" (one example was given in the previous section) are exceptions: they do not reveal an inconsistency or a total/partial redundancy that needs to be made explicit, since adding an instantiation is giving an example for a more general statement. It is important to reject an action introducing a redundancy instead of silently ignoring it because this often permits the author of the action to detect a mistake, a bad interpretation or a lack of precision (on his part or not). At the very least, this reminds the users that they should check what has already been represented on a subject before adding something on this subject.

4. *If the addition of a **new term u1#T** by a user u1 introduces an **inconsistency with statements of other users**, this **action is rejected by the system***. Indeed, such a conflict reveals that u1 has directly or indirectly used – and misunderstood - at least one term from another user in his definition of u1#T. The *addition by a user u2 of a definition to u1#T is actually a belief of u2* about the meaning of u1#T. This belief should be *rejected if it is found (logically) inconsistent with the definition(s) of u1#T by u1* (example in Point 6).

5. *If the addition, modification or removal of a statement **defining an already existing term u1#T** by a user u1 introduces an **inconsistency involving statements directly or indirectly re-using u1#T and created or believed by other users*** (i.e., users different from u1), **u1#T is automatically cloned** to solve this conflict and ensure that the original interpretation of u1#T by these other users is still represented. Indeed, such a conflict reveals that these other users had a more general interpretation of u1#T than u1 had or now has. Assuming that u2 is this other user or one of these other users, the term cloning of u1#T consists in creating u2#T with the same definitions as u1#T except for one, and then replacing u1#T by u2#T in the statements of u2. The difficulty is to chose a relevant definition to remove for the overall change of the KB to be minimal. In the case of term removal by u1, term cloning simply means changing the creator's identifier in this term to the identifier of one of the other users (if this generated term already exists, some suffix can be added). In a cbwoKB server, since statements point to the terms they use, changing an identifier does not require changing the statements. In a global virtual cbwoKB distributed on several servers, identifier changes in one server need to be replicated to other servers using this identifier. Manual term cloning is also used in knowledge integrations that are not loss-less (Djedidi and Aufaure, 2010).

    In a cbwoKB, it is not true that beliefs and term definitions "have to be updated sooner or later". To avoid this and to get precise knowledge, in a cbwoKB every belief must be contextualized in space and time, as in `u3#` `75% of bird can be agent of a flight´ in place France and in period 2005 to 2006´ (such contexts are not shown in the other examples of this article). If needed, u3 can associate the term `u3#75%_of_birds_fly__in_France_from_2005_to_2006` with this last belief. Due to the possibility of contextualizing beliefs, it is rarely necessary to create formal terms such as `u2#Sydney_in_2010`. Most common formal terms, e.g., `u3#bird` and `wordnet1.7#bird` never need to be modified by their creators. They are specializations of (or equal to) more general formal terms, e.g., `wn#bird` (the fuzzy concept of bird shared by all versions of the WordNet ontologies; `u3#bird` refers to a more precise concept, otherwise u3 would not have created it). What certainly evolves in time is the popularity of a belief or the popularity of the association between an informal term and a concept. If needed, this changing popularity can be represented by different statements contextualized in time and space.

6. *If adding, modifying or removing a **belief** introduces an **implicit potential conflict (partial/total inconsistency or redundancy) involving beliefs created by other creators**, it is **rejected***. However, a user may still represent his belief (say, b1) – and thus "loss-less correct" another user's belief that he does not believe in (say, b2) – by connecting b1 to b2 via a **corrective relation**. E.g., here are two FE statements by u2, each of which corrects a statement made earlier by u1:
    ```
    u2#` u1#`every bird is agent of a flight´ has for corrective_restriction
        u2#`most healthy flying_bird are able to be agent of a flight´ ´ and
    ```

```
u2#` u1#`every bird can be agent of a flight´ has for
    corrective_generalization u2#`75% of bird can be agent of a flight´ ´.
```

In the second case, u2's belief generalizes u1's belief and corrects it since otherwise u2 would not have needed to add it. In the first case, u2's belief specializes u1's belief (except for a quantifier which is generalized) and corrects it. In both cases, WebKB-2 detects the conflict by simple graph-matching.

If instead of the *belief* `every bird can be agent of a flight´ (all birds can fly), u1 entered the *definition* `any bird can be agent of a flight´, i.e., if he gave a *definition* to the type named "bird", there are two cases (as implied by the rules of the two previous points):

- u1 originally created this type (`u1#bird`); then, u2's attempt to correct the definition is rejected, or
- u1 added a definition to another user's type, say `wn#bird` since this WordNet type has no associated constraint preventing the adding of such a definition; then, i) the types `u1#bird` and `u2#bird` are automatically created as clones (and subtypes of) `wn#bird`, ii) the definition of u1 is automatically changed into `any u1#bird is agent of a flight´, and iii) the belief of u2 is automatically changed into `u2#`75% of u2#bird can be agent of a flight´.

In WebKB-2, users are encouraged to provide argumentation relations on corrective relations, i.e., a meta-statement using argument/objection relations on the statement using the corrective relation. However, to normalize the shared KB, people are encouraged not to use an objection relation but a "corrective relation with argument relations on them". Thus, not only are the objections stated but a correction is given and may be agreed with by several persons, including the author of the corrected statement (who may then remove it). Even more importantly, unlike objection relations, most corrective relations are transitive relations and hence their use permits better organization of argumentation structures, thus avoiding redundancies and easing information retrieval. The use of corrective relations makes explicit the disagreement of one user with (his interpretation of) the belief of another user. There is no inconsistency: an assertion A may be inconsistent with an assertion B but a belief that "A is a correction of B" is technically consistent with a belief in B. Thus, the shared KB can remain consistent.

For problem-solving purposes, application-dependent choices between contradictory beliefs often have to be made. To make them, an application designer can exploit i) the statements describing or evaluating the creators of the beliefs, ii) the corrective/argumentation and specialization relations between the beliefs, and more generally, iii) their evaluations via meta-statements (see Point 7). For example, an application designer may choose to select only the most specialized or restricted beliefs of knowledge providers having worked for more than 10 years in a certain domain. Thus, the approach of this protocol is unrelated to defeasible logics and avoids the problems associated with classic "version management" (furthermore, as above explained, in a cbwoKB, formal objects do not have to evolve in time).

This approach assumes that all beliefs can be argued against and hence be "corrected". This is true only in a certain sense. Indeed, among beliefs, one can distinguish "observations", "interpretations" ("deductions" or "assumptions"; in this approach, axioms are considered to be definitions) and "preferences"; although all these kinds of beliefs can be false (their authors can lie, make a mistake or assume a wrong fact), most people would be reluctant to argue against self-referencing beliefs such as `u2#"u2 likes flowers"` and `u2#"u2 is writing this sentence"`. The editing protocol of WebKB-2 relies on the reluctance of people to argue against such beliefs that should not be argued against.

7. To support more knowledge filtering or decision making possibilities and lead the users to be careful and precise in their contributions, a cbwoKB server should propose "default measures" deriving a global evaluation of each statement/creator from i) users' individual evaluations of these objects, and ii) global evaluations of these users. These measures should not be hard-coded but explicitly represented (and hence be executable) to let each user adapt them - i.e., combine their basic functions - according to his goals or preferences. Indeed, only the user knows the criteria (e.g., originality, popularity, acceptance, ..., number of arguments without objections on them) and weighting schemes that suit him. Then, since the results of these evaluations are also statements, they can be exploited by queries on the objects and/or their creators. Furthermore, before browsing or querying the cbwoKB, a user should be given the opportunity to set "filters for certain objects not to be displayed (or be displayed only in small fonts)". These filters may set conditions on statements about these objects or on the creators of these objects. They are automatically executed queries over the results of queries. In WebKB-2, filtering is based on a search for extended specialization, as for conceptual querying. Filters are useful when the user is overwhelmed by information in an insufficiently organized part of the KB. The KB server Co4 (Euzenat, 1996) also had protocols based on peer-reviewing for finding consensual knowledge; the result was a hierarchy of KBs, the uppermost ones containing the most consensual knowledge while the lowermost ones were the private KBs of contributing users. Establishing "how consensual a belief is" is more flexible in a cbwoKB: i) each user can design his own global measure for what it means to be consensual, and ii) KBs of consensual knowledge need not be generated. In any case, the reliability/popularity of user contributions is collaboratively assessed; this is much more difficult with traditional "static formal file based" approaches.

The approach described by the previous points is incremental and *works on semi-formal KBs*. Indeed, the users can set corrective or specialization relations between objects even when the system does not detect an inconsistency or redundancy. As noted, a new informal statement must be connected via an argumentation relation (e.g., a corrective relation) or an extended specialization relation to an already stored statement. For this relation to be correct, this new statement should generally not be composed of several sub-statements. However, allowing the storing of (small) paragraphs within a statement eases the incremental transformation of informal knowledge into (semi-)formal knowledge and allows doing so only when needed. This is necessary for the general acceptance of the approach. The techniques described in this article work do not seem particularly difficult for information technology amateurs, since the minimum they require is for the users to set the above mentioned relations from/to each term or statement. Hence, these techniques could be used in semantic wikis to avoid their governance problems cited in the introduction and other problems caused by their lack of structure. More generally, the presented approach removes or reduces the file-based approach problems listed in the previous section, without creating new problems. Its use would allow merging of (the information discussed or provided by the members of) many communities with similar interests, e.g., the numerous different communities working on the Semantic Web.

## 5.  EXAMPLES OF APPLICATIONS IN TEACHING

WebKB-2 has been used for integrating many ontologies (Martin, 2003, 2009) and representing many domains. In particular, it has been used for representing and inter-connecting the most important concepts of four different courses that I gave: "Workflow Management", "Systems Analysis & Design", "Introduction to Multimedia" and "Client-Server Architecture". Nearly each sentence of each slide for these courses has been represented into a semantic network of tasks, data structures, properties, definitions, etc. Figure 1 shows an extract of a Web file that was an input file for WebKB-2 and that mixed formal and informal elements; the formal ones are in the FL notation and represent important statements (here, relations between important concepts) from a book in Workflow Management. Figure 2 shows an example of results to a query. Each FL statement in these figures follow the generic schema:

```
CONCEPT1 RELATION1: CONCEPT2  CONCEPT3,
         RELATION2: CONCEPT4 (sourceForRel2) ...;
```

Such a statement should be read: "any CONCEPT1 may have for RELATION1 one or many CONCEPT2, and may have for RELATION1 one or many CONCEPT3, and may have for RELATION2 one or many CONCEPT4 (relation which can be found at sourceForRel2), ...". The sources of those relations in the book and the persons who created those representations (e.g., pm and the student s162557) are indicated. When the creator of a relation is not indicated, I (the user "pm") was the creator.

The students of these courses have recognised the help that the semantic network provided them in relating and comparing information otherwise scattered in many different slides and other lecture materials (an analysis of their evaluation of this teaching approach is given by Martin (2009)). However, having to learn the FL notation was perceived as a problem, especially by the students who were evaluated on their contributions to the semantic network. An intuitive table-based knowledge entering/display interface for FL should reduce this problem. Compared to an informal "learning journal", evaluating the students on their contributions to the semantic network permitted a much better evaluation of whether or not they understood the nature of the important concepts and their relationships. To enter these contributions, i.e., to collaboratively complete the initial "course formal summary (semantic network)" that I designed for them, the students used WebKB-2. For the students, the KB editing protocols were not a problem but entering meaningful knowledge representations proved to be very difficult. This highlighted the necessity for a very strong and very advanced semantic checking. Due to its knowledge normalization procedures, WebKB-2 enforces stronger semantic checks than

RDF+OWL inference engines but this still proved to be very insufficient.

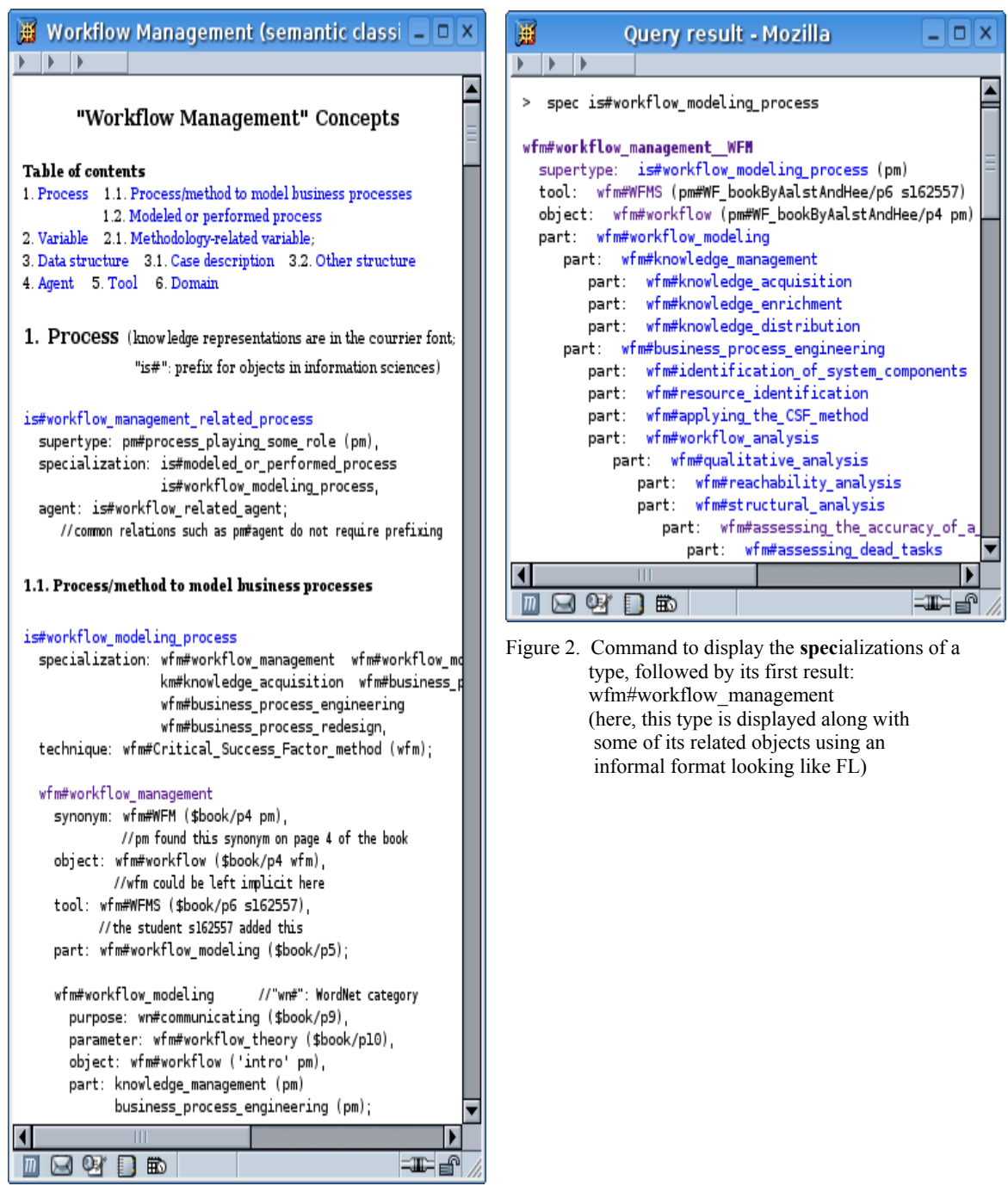

Figure 1. Extract from an input file including some
formal representations of representing statements
from a book in Workflow Management
(here referred to by the variable $book)

Figure 2. Command to display the **spec**ializations of a
type, followed by its first result:
wfm#workflow_management
(here, this type is displayed along with
some of its related objects using an
informal format looking like FL)

## 6. CONCLUSION

This article first aimed to show that a cbwoKB is *technically and socially* possible. To that end, the fourth (and main) section of this article presented a protocol permitting, enforcing or encouraging people to *incrementally* interconnect their knowledge into a well-organized (formal or *semi-formal*) KB without having to discuss and agree on terminology or beliefs. As noted, it seems that all other knowledge-based cooperation protocols that currently exists work on the comparison or integration of whole KBs, not on the comparison and loss-less integration of all their objects into a same KB. Other required elements for a cbwoKB - and for which WebKB-2 implements research results - were also mentioned: expressive and normalizing notations, methodological guidance, a large general ontology, and an initial cbwoKB core for the application domain of the intended cbwoKB. Already explored kinds of applications were cited. One currently explored is the collaborative representation and classification by Semantic Web experts of "Semantic Web related techniques". More generally, the approach seems interesting for collaboratively-built corporate memories or catalogues, e-learning, e-government, e-science, e-research, etc. Hillis (2004) describes a "Knowledge Web" to which teachers and researchers could add "isolated ideas" and "single explanations" at the right place, and suggests that this Knowledge Web could and should "include the mechanisms for credit assignment, usage tracking and annotation that the Web lacks" (pp. 4-5). Hillis did not give hints on what such mechanisms could be. The cbwoKB elements described by this article can be seen as a basis for such mechanisms.

A second aim of this article (mainly via Section 2) was to show that - in the long term or when creating a *new* KB for *general* knowledge sharing purposes - using a cbwoKB does/can provide more possibilities, with *on the whole* no more costs, than the mainstream approach (Shadbolt et al, 2006; Bizer et al., 2010) where knowledge creation and re-use involves searching, merging and creating (semi-)independent (relatively small) ontologies or semi-formal documents. The problem - and related debate - is more social (which formalism and methodology will people accept to learn and use?) than technical. A cbwoKB is much more likely to be adopted by a small communities of researchers but could incrementally grow to a larger and larger community. In any case, research on the two approaches are complementary: i) techniques of knowledge extraction or merging ease the creation of a cbwoKB, and ii) the results of applying these techniques with a cbwoKB as input would be better, and iii) these results would be easier to retrieve, compare, combine and re-use if they were stored in a cbwoKB.